\documentclass[letterpaper]{article} 
\usepackage{aaai2026}  
\usepackage{times}  
\usepackage{helvet}  
\usepackage{courier}  
\usepackage[hyphens]{url}  
\usepackage{graphicx} 
\usepackage{natbib}  
\usepackage{caption} 
\usepackage{algorithm}
\usepackage{algorithmic}

\usepackage{newfloat}
\usepackage{listings}
\DeclareCaptionStyle{ruled}{labelfont=normalfont,labelsep=colon,strut=off} 
\floatstyle{ruled}
\newfloat{listing}{tb}{lst}{}
\floatname{listing}{Listing}
\usepackage{booktabs}
\usepackage{amsfonts}
\usepackage{amsmath}
\usepackage{multirow}

\title{RacketVision: A Multiple Racket Sports Benchmark for Unified\\Ball and Racket Analysis}
\author{
    Linfeng Dong\textsuperscript{\rm 1,2}\thanks{Work done during internship at Shanghai AI Laboratory.}, 
    Yuchen Yang\textsuperscript{\rm 3,2}, 
    Hao Wu\textsuperscript{\rm 4,2}, 
    Wei Wang\textsuperscript{\rm 2}, 
    Yuenan Hou\textsuperscript{\rm 2}, \\
    Zhihang Zhong\textsuperscript{\rm 2}\thanks{Corresponding authors.}, 
    Xiao Sun\textsuperscript{\rm 2}\footnotemark[2]\\
}
\affiliations{
    \textsuperscript{\rm 1}Zhejiang University\\
    \textsuperscript{\rm 2}Shanghai AI Laboratory\\
    \textsuperscript{\rm 3}Fudan University\\
    \textsuperscript{\rm 4}University of Science and Technology of China\\

}

\usepackage{bibentry}

\begin{document}

\maketitle

\begin{abstract}
We introduce RacketVision, a novel dataset and benchmark for advancing computer vision in sports analytics, covering table tennis, tennis, and badminton. The dataset is the first to provide large-scale, fine-grained annotations for racket pose alongside traditional ball positions, enabling research into complex human-object interactions. It is designed to tackle three interconnected tasks: fine-grained ball tracking, articulated racket pose estimation, and predictive ball trajectory forecasting. Our evaluation of established baselines reveals a critical insight for multi-modal fusion: while naively concatenating racket pose features degrades performance, a Cross-Attention mechanism is essential to unlock their value, leading to trajectory prediction results that surpass strong unimodal baselines. RacketVision provides a versatile resource and a strong starting point for future research in dynamic object tracking, conditional motion forecasting, and multi-modal analysis in sports.
\end{abstract}


\section{Introduction}
\label{sec:introduction}

\begin{figure*}[htbp]
    \centering
    \includegraphics[width=0.95\linewidth]{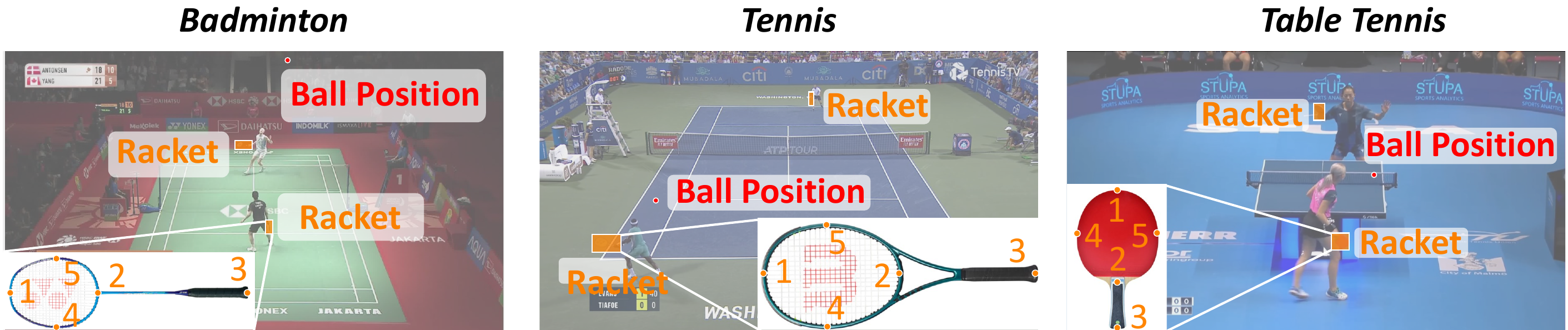}
    \caption{Visual examples of annotated data samples in RacketVision across the three sports. Each panel displays annotations for the ball's position (red dot) and the racket's bounding box (orange rectangle). The insets of each panel provide a schematic of the five keypoints defined for each specific racket type, which are used for the racket pose estimation task.}
    \label{fig:teasor}
\end{figure*}

Racket sports, typically represented by badminton, tennis, and table tennis, have garnered widespread global participation and attract research for performance analysis~\cite{kulkarni_table_2023, ttrobo2024, gossard_tt3d_2025}.
These sports encompass structurally defined computer vision tasks, while presenting perception challenges due to the rapid motion of both the ball and players, as well as the complex human-object interactions inherent in racket-based gameplay.
However, existing datasets~\cite{huang2019tracknet,sun2020tracknetv2,tarashima2023wasb} focus narrowly on ball tracking within a single sport at a time, falling short in two critical aspects:
1) They fail to leverage shared ball motion patterns across different sports.
2) Despite the racket being a central component, racket-specific annotations and analysis are lacking. This is crucial not only for sports analysis but also for complex neural avatar modeling~\cite{chen2024within,xu2025sequential,zhan2025r3,zhan2025towards}. These shortcomings limit the development of comprehensive racket sports analysis methods.

To address this gap, we introduce RacketVision, a multiple racket sports benchmark for unified ball and racket analysis.
RacketVision first expands the range of sports types for unified model training, aiming to uncover shared priors across racket sports.
Specifically, it comprises a collection of 1,672 video clips (435,179 frames, ~12,755 seconds) spanning badminton, tennis, and table tennis. 
In task design, RacketVision progressively proposes three tasks with corresponding annotations, enabling a more comprehensive decomposition of racket sport analysis.
Beyond the existing \textit{ball tracking} task, RacketVision defines racket keypoints and supports a novel \textit{racket pose estimation} task.
RacketVision further proposes an integrative task, \textit{ball trajectory prediction}, empowering downstream applications, such as tactic analysis~\cite{wang2024tacticai,SGA-INTERACT}, robotics~\cite{ttrobo2024,sci2025badmintonrobo}, etc.


In our evaluation, we establish extensive baselines for three tasks and analyze the impact of multi-sport training and multi-modal information under various fusion strategies. 
Our experiments reveal key insights that, training on all three sports generally enhances model generalization on perception tasks. More importantly, we uncover a nuanced relationship between multi-modal data and performance in trajectory prediction: a naive concatenation of racket pose features was found to be detrimental, performing worse than a ball-only baseline. However, by introducing a sophisticated Cross-Attention fusion mechanism, our LSTM-based model successfully leverages the racket information, ultimately outperforming the strong ball-only baseline across all three sports. This highlights that the value of racket pose data is critically dependent on the fusion architecture's ability to intelligently integrate contextual cues.

Our contributions are threefold:
\begin{itemize}
    \item We present RacketVision, a large-scale, multi-sport benchmark with detailed annotations for balls and rackets, supporting cross-sport analysis.
    \item We define three interconnected tasks, formulating key challenges of computer vision in sports analytics.
    \item We establish strong baseline solutions and conduct detailed evaluations, revealing key insights into multi-sport learning and the critical role of fusion architecture in multi-modal sports analysis.
\end{itemize}

\section{Related Work}
\label{sec:related_work}

\subsection{Racket Sport Datasets}
Existing datasets in racket sports have primarily focused on ball tracking. As summarized in Table~\ref{tab:dataset_compare}, while foundational datasets such as TrackNet~\cite{huang2019tracknet}, TrackNetv2~\cite{huang2019tracknet}, and WASB~\cite{tarashima2023wasb} enabled deep learning approaches for single-sport ball tracking, they presented opportunities for expansion in terms of scene diversity, sport variety, and annotation scope. Building upon these efforts, RacketVision provides a significant leap in scale and diversity, featuring substantially more games and frames across three distinct sports. Critically, it introduces the first large-scale annotations for racket pose (R) in addition to ball positions (B), enabling novel multi-modal analysis beyond simple ball tracking.

\begin{table}[th]
    \centering
    \setlength{\tabcolsep}{4pt}
    {
    \begin{tabular}{l|c|c|c|c|c|c}
    \toprule
    {Dataset} & {Res} & {\#S} & {\#G} & {\#F} & {\#A} & AT \\ 
    \midrule
    {GolfBall} & {$720p$} & 1 & {$1$} & {$2k$} & {$2k$} & B \\
    {OpenTTGames} & {$1080p$} & 1 & {$12$} & {$55k$} & {$55k$} & B \\
    {TrackNet} & {$720p$} & 1 & {$10$} & {$19k$} & {$19k$} & B \\
    {TrackNetv2} & {$720p$} & 1 & {$19$} & {$78k$} & {$78k$} & B \\
    \midrule
    {RacketVision} & {$1080p$} & 3 & {$461$} & {$435k$} & {$88k$} & B,R\\
    \bottomrule
    \end{tabular}
    \caption{Comparison of racket sports datasets. Res stands for resolution. \#G, \#F and \#A stands for number of Games, Frames and Annotations. \#S stands for number of sport types. AT stands for annotation types, where B is ball position, R is our first proposed racket pose annotation. }
    \label{tab:dataset_compare}}
\end{table}

\subsection{Racket Sport Analysis Methods}
Research on sport analysis methods has evolved from the basic task of 2D ball tracking to more sophisticated tasks centered on humans and games~\cite{xia2024sportu, rao2024unisoccer, dong2024lucidaction, yang2024X, LearnableSMPLify}. The development of robust 2D trackers has progressed from early CNN-based detectors~\cite{reno_convolutional_2018}, to specialized architectures for small objects~\cite{jedrzejczak_deepball_2019}, and data-efficient semi-supervised learning~\cite{vandeghen_semi-supervised_2022}. The importance of this task is further underscored by large-scale benchmarks like SoccerNet~\cite{Deliege2020SoccerNetv2, cioppa_soccernet-tracking_2022}. Recent studies have explored 3D trajectory and spin reconstruction from monocular videos~\cite{gossard_tt3d_2025, kienzle_towards_2025}, hit anticipation~\cite{etaat_latte-mv_2025}, and stroke recognition using trajectory data alone~\cite{kulkarni_table_2023}. Racket-centric studies rely on specialized hardware like high-speed or stereo cameras~\cite{chen_visual_2013, gao_real-time_2019} or complex proxies like human keypoints to handle occlusions~\cite{zheng_method_2023}. However, these advanced methods have been constrained by the lack of large-scale, unified benchmarks. RacketVision addresses this gap, providing a public benchmark to train and evaluate general-purpose models~\cite{jiang2023rtmpose, xu2022vitpose, carion2020end, cao2017realtime, yolo11_ultralytics} on these complex, interconnected analysis tasks, thereby lowering the barrier for future research.

\section{RacketVision Dataset}
\label{sec:dataset_description}

We introduce RacketVision, a large-scale video dataset designed to foster research in sports analytics across multiple racket sports: table tennis, tennis, and badminton. The benchmark provides a comprehensive resource for the interconnected tasks of ball tracking, racket pose estimation, and trajectory prediction. 
A detailed statistical breakdown for each sport is provided in Table~\ref{tab:dataset_statistics}.

\subsection{Data Collection and Annotation}
\label{subsec:pipeline}

\begin{figure}[htbp]
    \centering
    \includegraphics[width=\linewidth]{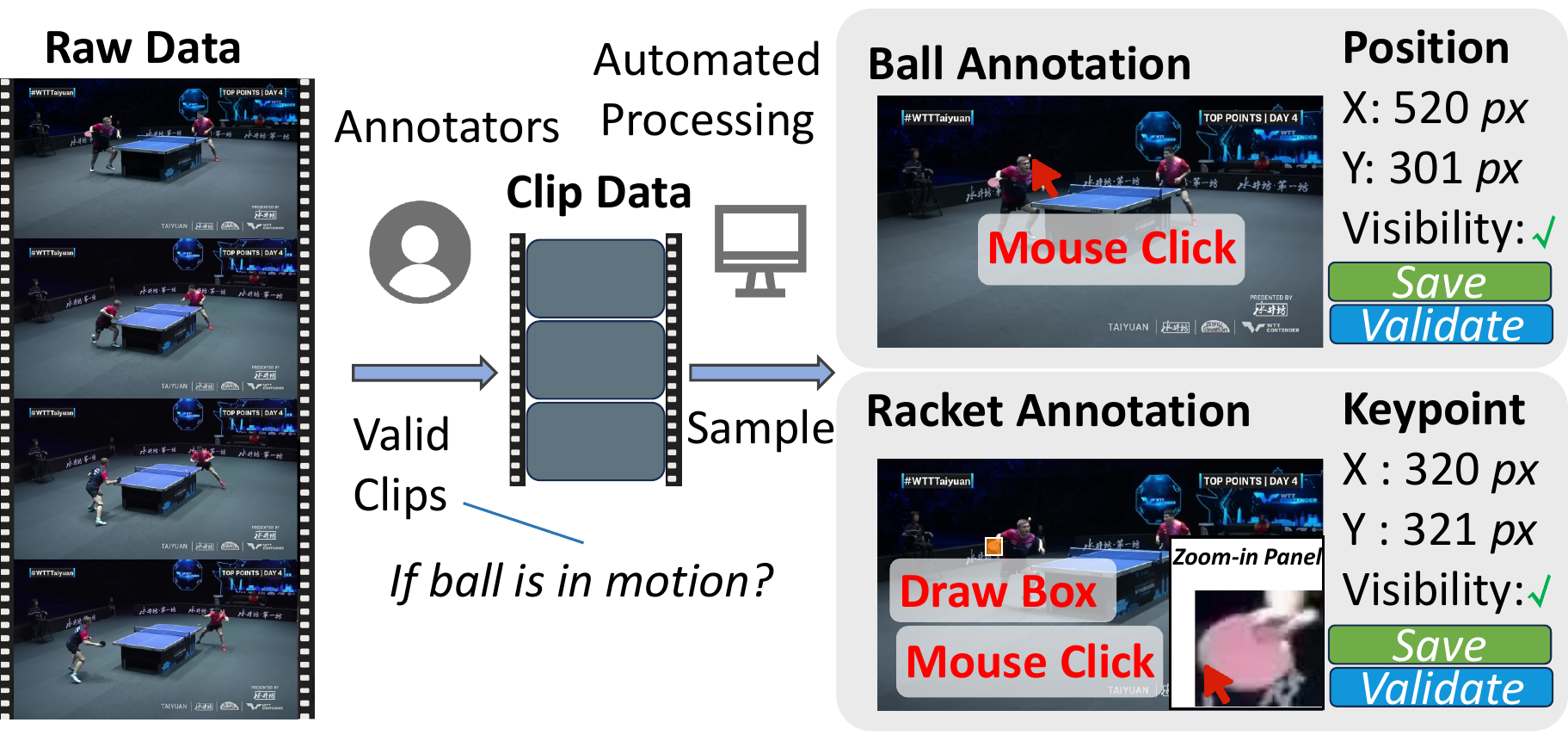}
    \caption{The two-stage annotation pipeline for RacketVision. First, crowd-sourced annotators segment valid clips from raw videos where the ball is in motion. Second, on sparsely sampled frames from these clips, another group of annotators labels the ball's position as well as the racket's bounding box and keypoints using a specialized interface.}
    \label{fig:annot}
\end{figure}

The data collection and annotation pipeline, illustrated in Figure~\ref{fig:annot}, was designed to ensure data quality, diversity, and annotation efficiency. The process begins with sourcing video from 942 top-level professional game broadcasts on YouTube, covering badminton, tennis, and table tennis to capture a wide variety of players and match dynamics. 

In the first stage of the pipeline, a team of crowd-sourced annotators segments these raw videos into valid clips. A clip is defined as a continuous segment of 5-10 seconds where the ball is actively in play, which focuses the dataset on the most analytically relevant portions of the game. In the second stage, we employ a sparse annotation strategy to balance cost and quality. Instead of annotating every frame, 20\% of the frames within each clip are evenly sampled for manual labeling by a different group of annotators. This approach maintains high temporal diversity while reducing redundancy. As illustrated in Figure~\ref{fig:teasor}, annotators labeled the ball's position as a single point (red dot) with a visibility flag for each sampled frame. Rackets were annotated with both a bounding box (orange rectangle) and five specific keypoints designed to capture the pose of each racket type.

\begin{table}[t]
  \centering
  \setlength{\tabcolsep}{2pt}
  \resizebox{.46\textwidth}{!}{
  \begin{tabular}{l|cccccc}
    \toprule
    Sport & {\#Games} & {\#Clips} & {\#Frames} & {\shortstack{Length~(s)}} & {\shortstack{Ball \\ Anno.}} & {\shortstack{Racket \\ Anno.}} \\
    \midrule
    Table Tennis & 50 & 780 & 170,027 & 3,878 & 19,495 & 6,648 \\
    Tennis       & 431 & 431 & 150,399 & 4,285 & 21,544 & 7,395 \\
    Badminton    & 461 & 461 & 114,753 & 4,592 & 23,003 & 10,578 \\
    \midrule
    Total        & 942 & 1672 & 435,179 & 12,755 & 64,042 & 24,621 \\
    \bottomrule
  \end{tabular}}
    \caption{Statistical breakdown of the RacketVision dataset, detailing the number of games, clips, frames, total duration (in seconds), and the count of ball and racket annotations for each sport and in total.}
    \label{tab:dataset_statistics}
\end{table}

\subsection{Dataset Structure}
\label{subsec:data_structure}

Each sample in RacketVision is a short video clip accompanied by a metadata file. The metadata includes the sport type and the indices of the annotated frames. To support baseline models that leverage background information, we also pre-process and provide a median frame for each clip. This serves as a stable background reference, which is particularly useful for distinguishing the small, fast-moving ball from the environment. All annotations, including ball positions, racket bounding boxes, and racket keypoints, are provided at the pixel level.

\section{Tasks}
\label{sec:tasks}

\begin{figure*}
    \centering
    \includegraphics[width=\linewidth]{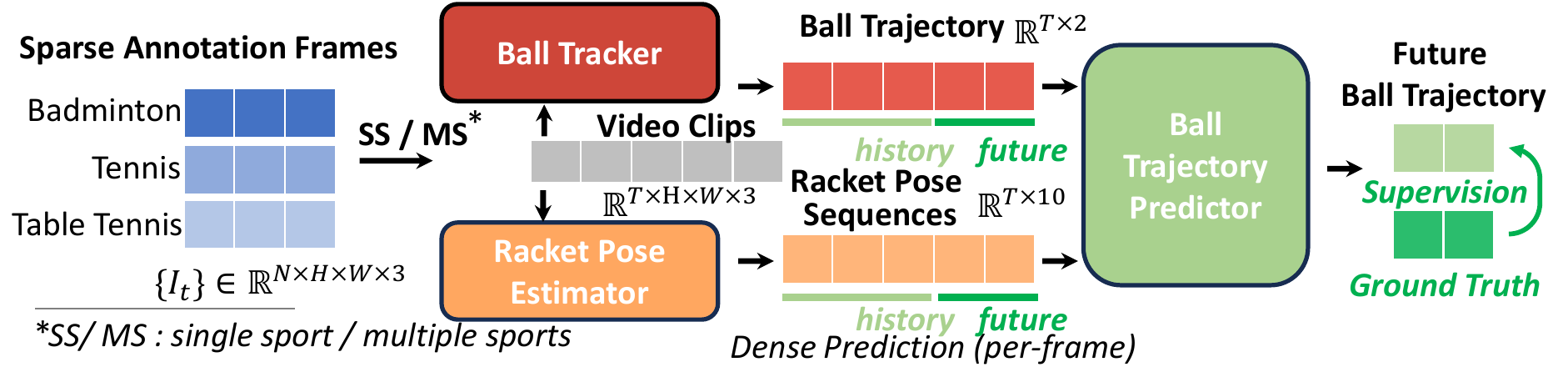}
  \caption{An overview of the task pipeline in RacketVision. Initially, the \textit{Ball Tracker} and \textit{Racket Pose Estimator} are trained using sparse ground-truth annotations. These models then process full video clips to generate dense trajectory data (soft labels), which serves as the training input for the final \textit{Ball Trajectory Predictor}.}
  \label{fig:task_pipeline}
\end{figure*}

The RacketVision benchmark is structured around three interconnected tasks that form a comprehensive pipeline for sports analysis, progressing from low-level perception to high-level prediction: ball tracking, racket pose estimation, and ball trajectory prediction. Together, they serve a dual purpose: to drive innovation in sports analytics and to provide a framework for studying multi-modal, dynamic human-object interactions. 

Figure~\ref{fig:task_pipeline} provides a detailed schematic of the relationship and workflow between these tasks. As illustrated, the process begins with training the two foundational perception models, the Ball Tracker and the Racket Pose Estimator, directly on the sparse, manually-labeled ground-truth frames provided in our dataset. Subsequently, these trained perception models are deployed on full video clips to generate dense, per-frame predictions, or "soft labels," of ball and racket positions. These continuous sequences are then segmented into historical and future data segments, forming the rich training dataset for the final high-level task: the Ball Trajectory Predictor. This pipeline structure not only defines the dependencies between the tasks but also represents a practical workflow for building a complete sports analysis system. The following sections will provide a formal problem definition for each task.

\subsection{Ball Tracking}
\label{subsec:ball_tracking}

\textbf{Problem Formulation.} Given RGB frame \( I_t \in \mathbb{R}^{H \times W \times 3} \) at time \( t \) (single-frame setting) or a sequence of frames \( \{I_{t-N}, \ldots, I_t\} \) with \( N=5 \) (multi-frame setting), predict the ball’s position \( (x_t, y_t) \in \mathbb{R}^2 \) and visibility flag \( v_t \in \{0, 1\} \) in the target frame \( I_t \).

\noindent
\textbf{Settings.} The task has two settings: (1) \textit{single-frame}, using only the target frame \( I_t \) to test static detection capabilities; and (2) \textit{multi-frame}, using the target frame and five preceding frames \( \{I_{t-5}, \ldots, I_t\} \) to leverages temporal context for improved robustness against occlusions and motion blur~\cite{rozumnyi2021shape,zhong2022animation,zhong2023blur}.

\subsection{Racket Pose Estimation}
\label{subsec:racket_pose_estimation}

\textbf{Problem Formulation.} Given RGB frame \( I_t \in \mathbb{R}^{H \times W \times 3} \), predict the bounding box \( (x_{\text{min}}, y_{\text{min}}, x_{\text{max}}, y_{\text{max}}) \in \mathbb{R}^4 \) and five keypoints \( \{(x_i, y_i)\}_{i=1}^5 \in \mathbb{R}^{10} \) for each racket in the frame.

\noindent
\textbf{Settings.} The task uses a single-frame setting, predicting Bbox and keypoints from \( I_t \). This setting focuses on static pose estimation, suitable for the dataset’s high-resolution frames and diverse racket orientations.

\subsection{Ball Trajectory Prediction Given History}
\label{subsec:trajectory_prediction}

\textbf{Problem Formulation.} Given a history of ball positions over \( N \) frames \( \{(x_{t-N+1}, y_{t-N+1}), \ldots, (x_t, y_t)\} \in \mathbb{R}^{N \times 2} \) (ball-only setting) or ball positions plus racket poses \( \{(x_{t-N+1}, y_{t-N+1}, \\ \{(x_i, y_i)\}_{i=1}^5), \ldots, (x_t, y_t, \{(x_i, y_i)\}_{i=1}^5)\} \) (ball + racket setting), predict the ball’s trajectory over the next \( M \) frames \( \{(\hat{x}_{t+1}, \hat{y}_{t+1}), \ldots\\, (\hat{x}_{t+M}, \hat{y}_{t+M})\} \in \mathbb{R}^{M \times 2} \). 

\noindent
\textbf{Settings.} The task has two settings for input data modality: (1) \textit{ball-only}, using the \( N \)-frame ball position history, which tests trajectory modeling based on ball dynamics; and (2) \textit{ball + racket}, incorporating racket pose history (5 keypoints), which accounts for player interactions and improves prediction accuracy. We also use two settings for (1) \textit{long trajectory}, that set \( N=80 \) and \( M=20 \), and (2) \textit{short trajectory}, that set \( N=20 \) and \( M=5 \). 

\section{Experiments and Baseline Solutions}
\label{sec:experiments}

We evaluate the RacketVision dataset on tasks defined in Sec.\ref{sec:tasks}: ball tracking, racket pose estimation, and ball trajectory prediction given history. For each task, we define evaluation metrics, introduce baseline models, present experimental results, and analyze key observations. Figure.~\ref{fig:task_pipeline} shows the relationships between the 3 tasks. 

\subsection{Ball Tracking}
\label{subsec:ball_tracking_exp}

\textbf{Evaluation Metrics.} We use four standard metrics to evaluate ball tracking performance:
\begin{itemize}
    \item \textit{Precision (Prec.)}: The ratio of correctly predicted ball positions (within a distance threshold) to the total number of predictions.
    \item \textit{Recall (Rec.)}: The ratio of correctly predicted ball positions to the total number of ground-truth visible balls.
    \item \textit{Mean Distance Error (MDE)}: The average Euclidean distance in pixels between predicted and ground-truth positions for visible balls, assuming a 1920x1080 resolution. A lower value is better.
    \item \textit{mAP@50 (mAP)}: Mean Average Precision at IoU threshold of 0.5, evaluating the overall detection accuracy.
\end{itemize}

\noindent
\textbf{Baseline Models.} We evaluate three representative baseline models:
\begin{itemize}
    \item \textit{RTMDet}~\cite{lyu2022rtmdet}: A state-of-the-art real-time object detection model, adapted for ball detection.
    \item \textit{YOLO11}~\cite{yolo11_ultralytics}: A state-of-the-art vision model for real-time object detection. 
    \item \textit{WASB}~\cite{tarashima2023wasb}: A strong baseline specifically designed for sports ball tracking, which internally incorporates background modeling.
    \item \textit{TrackNetV3}~\cite{chen2023tracknetv3}: A specialized heatmap-based network for tracking small, high-speed objects in sports, which can leverage temporal context from multiple frames.
\end{itemize}

\noindent
\textbf{Experimental Setup.} Our experiments are designed to investigate three key axes of performance: the choice of model architecture, the benefit of multi-sport training, and the impact of techniques like background modeling (BM) and multi-frame inputs (\#F). Due to the extensive search space, we focused our multi-frame and multi-sport experiments primarily on the \textit{TrackNetV3} architecture. 

\begin{figure}[htbp]
    \centering
    \includegraphics[width=\linewidth]{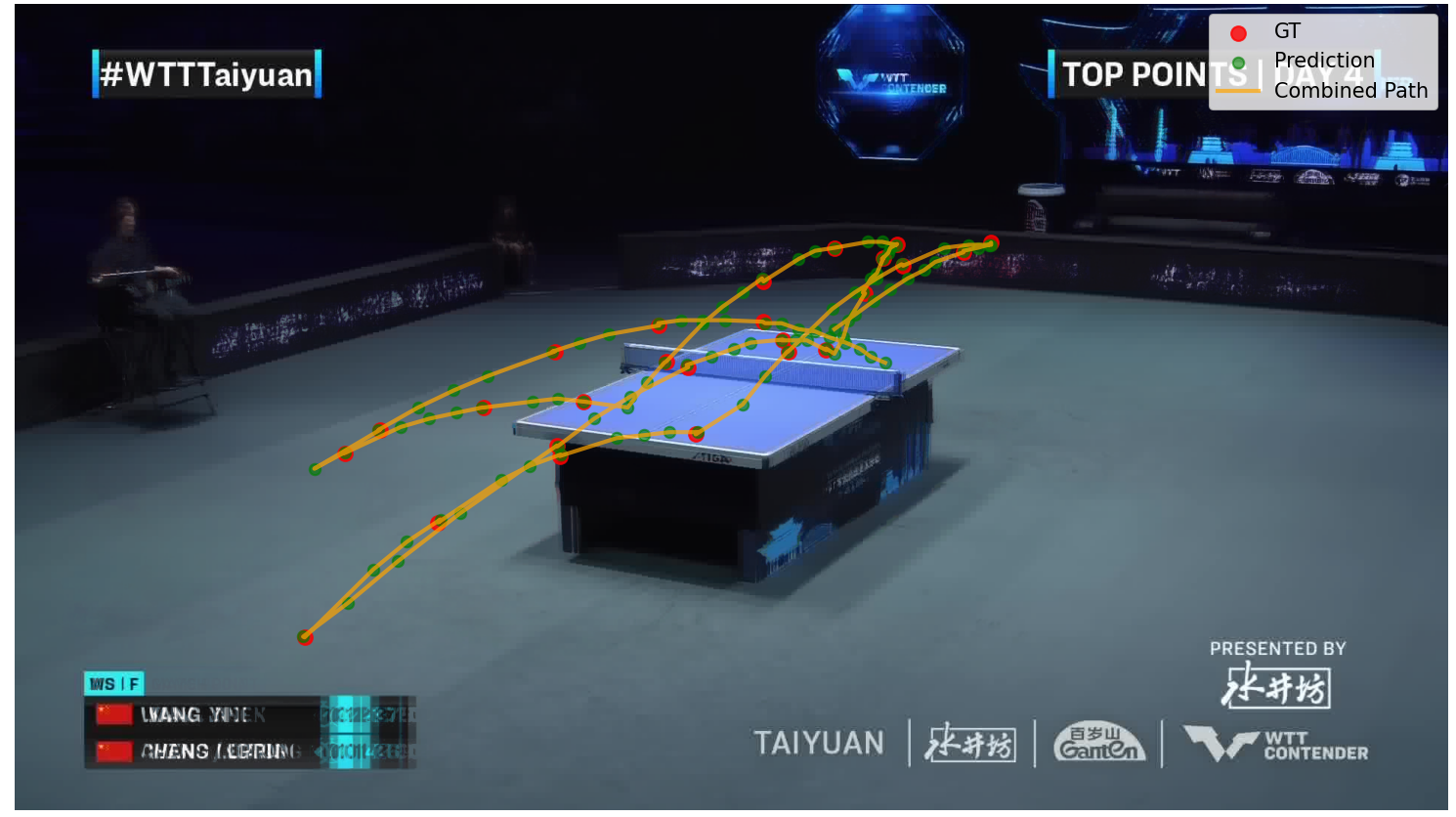}
    \caption{The visualization of ball tracking result of MS-TrackNetV3~(with BM, \#F=4) on table tennis. The red dots are sparse ground-truth ball position annotations, while the green dots are model predictions. The yellow line shows the combined path of ground-truth and predictions, illustrating the complete ball trajectory within the clip.}
    \label{fig:vis1}
\end{figure}

\noindent
\textbf{Experimental Results.} Table~\ref{tab:ball_tracking_results} summarizes the performance of all evaluated models and settings. Figure~\ref{fig:vis1} provides a visual example of the tracking performance of our best model.

\begin{table*}[htbp]
  \centering
  \setlength{\tabcolsep}{4pt}
  \begin{tabular}{@{}l c c cccc cccc cccc@{}}
    \toprule
    \multirow{2}{*}{Model} & \multirow{2}{*}{\shortstack{BM}} & \multirow{2}{*}{\shortstack{\#F}} & \multicolumn{4}{c}{Table Tennis} & \multicolumn{4}{c}{Tennis} & \multicolumn{4}{c}{Badminton} \\
    \cmidrule(lr){4-7} \cmidrule(lr){8-11} \cmidrule(lr){12-15}
    & & & Prec. & Rec. & MDE & mAP & Prec. & Rec. & MDE & mAP & Prec. & Rec. & MDE & mAP \\
    \midrule
    RTMDet     & $\times$ & 1 & 0.844 & 0.724 & 35.8 & 68.1 & 0.862 & 0.477 & 22.5 & 46.7 & 0.801 & 0.721 & 38.2 & 64.2 \\
    YOLO11 & $\times$ & 1 & 0.877 & 0.595 & 30.6 & 67.2 & 0.881 & 0.416 & 17.5 & 58.5 & 0.858 & 0.548 & 29.3 & 72.1 \\ 
    WASB             & \checkmark & 1 & 0.843 & \underline{0.733} & 12.9 & 51.8 & 0.937 & 0.803 & 3.62  & 66.0 & 0.907 & 0.803 & 4.03  & 58.7 \\
    TrackNetV3       & $\times$ & 1 & 0.793 & 0.639 & 22.5 & 51.0 & 0.904 & 0.725 & 7.95  & 58.9 & 0.876 & 0.706 & 7.39  & 57.8 \\
    TrackNetV3       & \checkmark & 1 & 0.869 & 0.724 & 10.3 & 67.7 & 0.941 & \underline{0.830} & 3.07  & 65.6 & \underline{0.924} & 0.797 & 3.34  & 67.7 \\
    TrackNetV3       & \checkmark & 4 & \underline{0.898} & 0.725 & \underline{6.63}  & \underline{68.3} & \underline{0.962} & 0.797 & \underline{1.66}  & \underline{68.7} & 0.922 & \underline{0.865} & \underline{2.07}  & \underline{72.5} \\
    \midrule
    MS-RTMDet     & $\times$ & 1 & 0.852 & 0.733 & 29.6 & 69.3 & 0.885 & 0.503 & 19.3 & 48.3 & 0.823 & 0.736 & 31.0 & 64.5 \\
    MS-WASB           & \checkmark & 1 & 0.867 & 0.742 & 8.53  & 56.2 & 0.943 & 0.802 & 3.73  & 65.1 & 0.924 & 0.807 & 2.13  & 63.3 \\
    MS-TrackNetV3     & $\times$ & 1 & 0.804 & 0.672 & 19.6  & 52.4 & 0.912 & 0.769 & 6.58  & 50.2 & 0.830 & 0.727 & 6.95  & 59.4 \\
    MS-TrackNetV3     & \checkmark & 1 & 0.890 & 0.731 & 7.57  & 59.1 & \textbf{0.962} & 0.820 & \textbf{1.70}  & 66.6 & 0.906 & 0.824 & 3.68  & 67.5 \\
    MS-TrackNetV3     & \checkmark & 4 & \textbf{0.924} & \textbf{0.762} & \textbf{3.41}  & \textbf{71.1} & 0.945 & \textbf{0.880} & 1.96  & \textbf{81.9} & \textbf{0.915 }& \textbf{0.865} & \textbf{1.54}  & \textbf{83.1} \\
    \bottomrule
  \end{tabular}
  \caption{Ball Tracking Experimental Results on RacketVision. BM represents whether add background median into input. Models starts with MS- are trained on all three sports, while others are trained on one sport. The bold results are the best results of MS-models, the underline results are the best results of models trained on single sport.}
  \label{tab:ball_tracking_results}
\end{table*}

\noindent
\textbf{Observations and Analysis.} Table~\ref{tab:ball_tracking_results} summarizes the performance of all evaluated models and settings. Figure~\ref{fig:vis1} provides a visual example of the tracking performance of our best model.

\begin{itemize}

    \item \textbf{Multi-Sport training generally enhances model generalization, especially on detection-oriented metrics.} By comparing the best multi-sport model (MS-TrackNetV3, \#F=4, bold results) with the best single-sport model (TrackNetV3, \#F=4, underlined results), we observe a clear trend of improved generalization. For example, the MS model boosts mAP by a significant 19.2\% in Tennis (81.9 vs. 68.7) and 14.6\% in Badminton (83.1 vs. 72.5). However, this broad generalization sometimes comes at the cost of hyper-specialized precision on a single sport. For instance, the single-sport model retains a slight edge in Tennis precision (0.962 vs. 0.945) and MDE (1.66 vs. 1.96). This suggests that training on diverse data forces the model to learn more robust features, enhancing its ability to find the ball under varied conditions (higher Recall and mAP), occasionally at the expense of pinpoint localization accuracy on a specific domain.

    \item \textbf{Background modeling is a highly effective technique for reducing localization error.} Incorporating a median frame for background subtraction (BM=$\checkmark$) provides a powerful prior for distinguishing the small, fast-moving ball from a static or complex background. This is most evident in the Mean Distance Error (MDE). For example, comparing TrackNetV3 (\#F=1) with and without BM, background modeling reduces MDE by a remarkable 54.0\% for Table Tennis, 61.4\% for Tennis, and 54.8\% for Badminton. This consistently large improvement underscores the value of providing the model with an explicit representation of the static scene to mitigate false positives and improve localization.

    \item \textbf{Leveraging temporal context with multi-frame inputs boosts overall detection accuracy but reveals performance trade-offs.} Using multiple frames (\#F=4) allows the model to leverage motion cues, which is particularly effective for improving recall and mAP in complex scenarios. The best overall results in our benchmark are achieved by MS-TrackNetV3 with 4 frames. However, the claim that multi-frame input is universally superior requires nuance. For instance, while using 4 frames with MS-TrackNetV3 in Tennis boosts Recall (0.880 vs. 0.820), the single-frame version achieves a slightly better MDE (1.70 vs. 1.96). This indicates that while temporal context is crucial for detecting the ball during challenging rallies (improving mAP), it can occasionally introduce minor jitter or motion blur that slightly affects the precision of the final predicted coordinate compared to a clean single frame.
\end{itemize}

\subsection{Racket Pose Estimation}
\label{subsec:racket_pose_estimation_exp}

\noindent
\textbf{Evaluation Metrics.} We use two metrics to evaluate racket pose estimation:
\begin{itemize}
    \item \textit{Percentage of Correct Keypoints@0.2~(PCK)}: Percentage of predicted keypoints that fall within a normalized distance threshold of 0.2 times the bounding box size from their corresponding ground-truth positions. 
    
    \item \textit{Mean Per-Joint Position Error~(MPJPE)}: Average Euclidean distance in pixels between predicted and ground-truth keypoint positions across all visible keypoints. 
    
    \item \textit{mean Object Keypoint Similarity~(mOKS)}: Mean Object Keypoint Similarity score that measures the similarity between predicted and ground-truth keypoint configurations.
    
    \item \textit{Normalized Mean Error~(NME)}: Normalized mean error calculated by dividing the average keypoint position error by the distance between left and right keypoints, providing scale-invariant evaluation of pose estimation accuracy.
    
    \item \textit{{mAP@50~(mAP)}: Mean Average Precision at IoU=0.5, evaluating detection accuracy.}: Average precision computed at Intersection over Union (IoU) threshold of 0.5, measuring detection accuracy for moderately overlapping predictions with ground-truth bounding boxes.
\end{itemize}

\noindent
\textbf{Baseline Models.} We evaluate a top-down baseline in single-sport and multi-sport training settings:
\begin{itemize}
    \item \textit{RTMPose}~\cite{jiang2023rtmpose}: A real-time top-down pose estimation model, optimized for keypoint detection in single-frame inputs. We adopt RTMDet~\cite{lyu2022rtmdet} as detector to generate bounding box.
\end{itemize}

\begin{figure}[htbp]
    \centering
    \includegraphics[width=\linewidth]{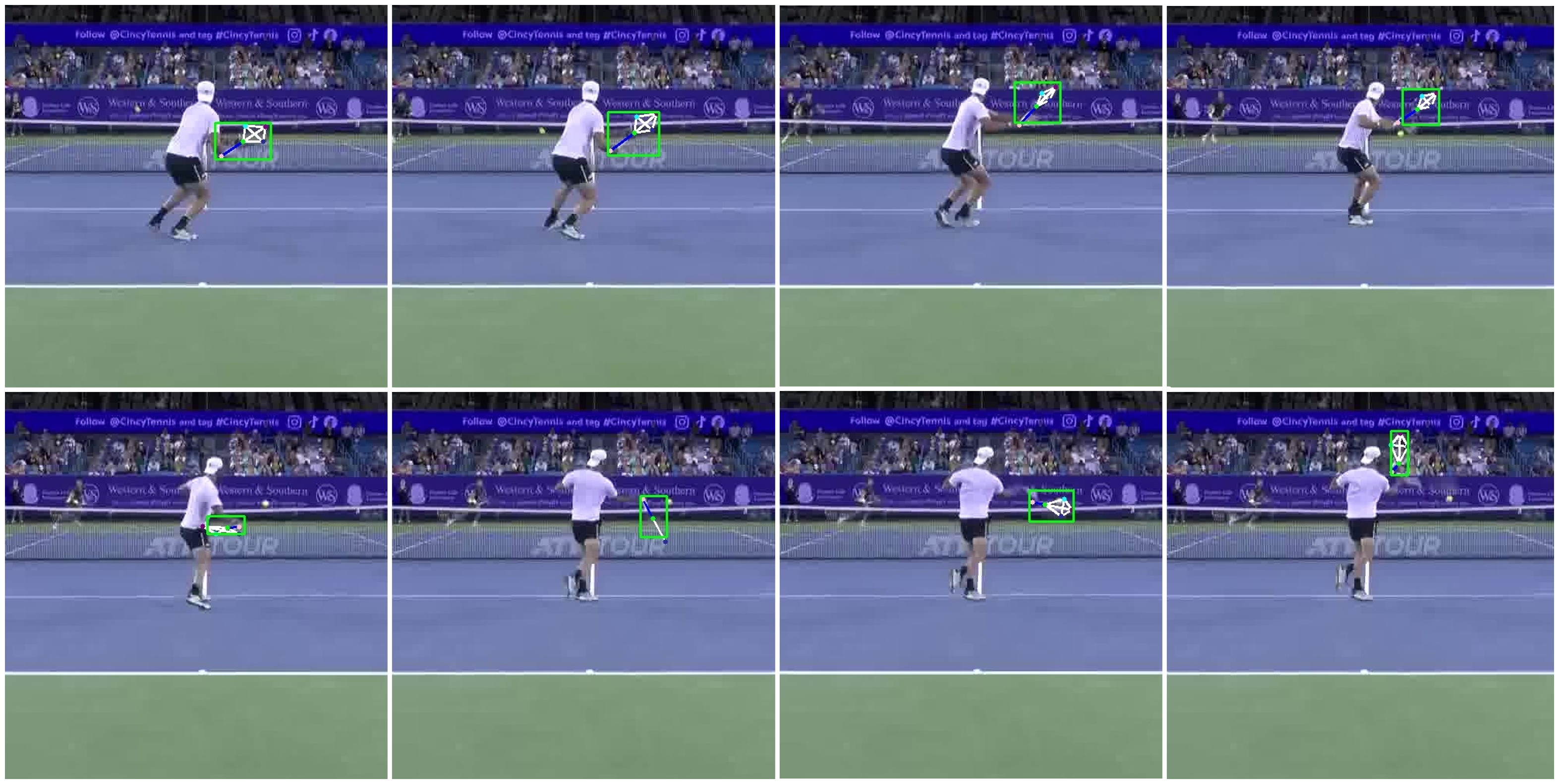}
    \caption{Visualization result of racket pose estimation of MS RTMPose model on tennis clip.}
    \label{fig:vis2}
\end{figure}

\noindent
\textbf{Experimental Results.} Table~\ref{tab:racket_pose_results} summarizes the performance of baselines under the single-frame setting. Table~\ref{tab:racket_keypoint_analysis} compares the performance of the best model on 5 keypoints. Figure~\ref{fig:vis2} provides a visual example of the racket pose estimation result of our best model.

\begin{table}[htbp]
\centering
\setlength{\tabcolsep}{4pt}
\resizebox{.46\textwidth}{!}{
\begin{tabular}{l|l|cccc|c}
\toprule
\multirow{2}{*}{\textbf{Train}} & \multirow{2}{*}{\textbf{Sport}} & \multicolumn{4}{c|}{\textbf{Pose Estimation}} & \textbf{Det} \\
\cmidrule(lr){3-7}
& & \textbf{PCK} & \textbf{MPJPE} & \textbf{mOKS} & \textbf{NME} & \textbf{mAP} \\
\midrule
\multirow{3}{*}{\textbf{SS}} & Table Tennis & 75.6 & 10.6 & 0.453 & 0.279 & 72.4 \\
& Tennis & 83.7 & 5.87 & 0.574 & 0.245 & 73.4 \\
& Badminton & 82.1 & 5.45 & 0.601 & 0.259 & 69.8 \\
\midrule
\multirow{3}{*}{\textbf{MS}} & Table Tennis & 81.8 & 9.71 & 0.498 & 0.254 & 78.4 \\
& Tennis & \textbf{89.6} & 5.34 & 0.630 & \textbf{0.223} & \textbf{79.4} \\
& Badminton & 88.5 & \textbf{5.00} & \textbf{0.668} & 0.235 & 75.5 \\
\bottomrule
\end{tabular}}
\caption{Main Performance Metrics Comparison. MS representing model trained on multiple sports, while SS representing single sport.}
\label{tab:racket_pose_results}
\end{table}

\begin{table}[htbp]
\centering
\setlength{\tabcolsep}{4pt}
\begin{tabular}{l|ccccc}
\toprule
\multirow{2}{*}{\textbf{Sport}} & \multicolumn{5}{c}{\textbf{Individual Keypoints (\%)}} \\
\cmidrule(lr){2-6}
& \textbf{Top} & \textbf{Bottom} & \textbf{Handle} & \textbf{Left} & \textbf{Right} \\
\midrule
Table Tennis & 97.6 & 97.3 & \textbf{97.9} & 64.8 & 64.8 \\
Tennis & 98.6 & 98.9 & 92.6 & \textbf{79.7} & \textbf{80.1} \\
Badminton & \textbf{99.4} & \textbf{99.7} & 97.3 & 74.6 & 75.5 \\
\bottomrule
\end{tabular}
\caption{The comparison of PCK@0.2 performance of MS RTMPose model on different racket pose keypoints.}
\label{tab:racket_keypoint_analysis}
\end{table}

\begin{figure}[htbp]
  \centering
  \includegraphics[width=0.95\linewidth]{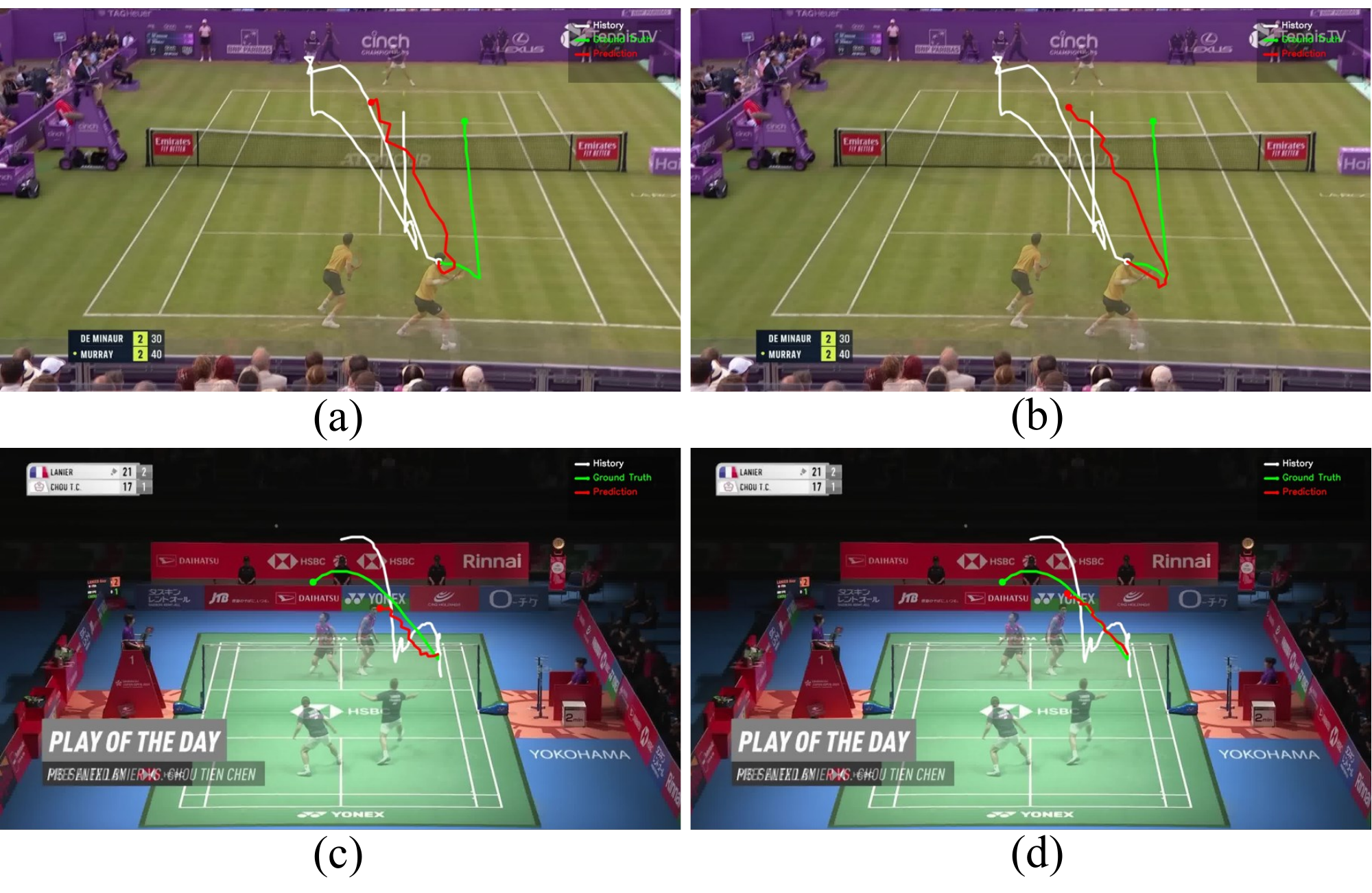}
  \caption{Qualitative comparison of long trajectory prediction. We compare the baseline LSTM Ball-Only model in (a), (c) with our proposed Cross-Attention LSTM Ball+Racket model in (b), (d).}
  \label{fig:traj_pred_compare}
\end{figure}
\begin{itemize}
    \item \textit{Multi-sport training outperforms single-sport.} The MS model achieves superior performance with PCK@0.2 improvements of 6.17\%, 6.36\%, and 5.97\% for table tennis, badminton, and tennis respectively. Tennis reaches the highest overall PCK@0.2 of 89.69\% under multi-sport training, while badminton excels in pose quality metrics with the lowest MPJPE (5.00px) and highest mOKS (0.668), demonstrating the effectiveness of cross-sport knowledge transfer.
    \item \textit{Side keypoint detection poses significant challenges for accurate racket pose estimation.} While structural keypoints (top, bottom, handle) achieve excellent accuracy above 92\%, side keypoints (left, right) exhibit substantially lower performance ranging from 64.85\% to 80.11\%. This disparity stems from the inherent difficulty of detecting side edges, which are often occluded by hand grip, subject to motion blur during rapid movements, and highly sensitive to viewing angles. 
\end{itemize}

\subsection{Ball Trajectory Prediction Given History}
\label{subsec:trajectory_prediction_exp}

\begin{table*}[htbp]
  \centering
  \setlength{\tabcolsep}{8pt}
  \begin{tabular}{@{}l l l l rr rr rr@{}}
    \toprule
    \multirow{2}{*}{Setting} & \multirow{2}{*}{Model} & \multirow{2}{*}{Input} & \multirow{2}{*}{Fuse Method} & \multicolumn{2}{c}{Table Tennis} & \multicolumn{2}{c}{Tennis} & \multicolumn{2}{c}{Badminton} \\
    \cmidrule(lr){5-6} \cmidrule(lr){7-8} \cmidrule(lr){9-10}
    & & & & ADE & FDE & ADE & FDE & ADE & FDE \\
    \midrule
    \multicolumn{10}{l}{\textit{Short Trajectory Prediction (History=20, Future=5)}} \\
    \midrule
    Short & LSTM & Ball & - & 41.9 & 64.0 & 23.8 & 37.6 & 37.5 & 60.7 \\
    Short & LSTM & Ball+Racket & Concat & 58.1 & 86.6 & 29.3 & 45.3 & 45.7 & 70.7 \\
    Short & LSTM & Ball+Racket & CrossAttn & \textbf{38.3} & \textbf{60.4} & \textbf{22.8} & \textbf{35.7} & \textbf{37.0} & \textbf{59.3} \\
    Short & Transformer & Ball & - & 46.9 & 67.8 & 35.7 & 47.3 & 43.9 & 64.8 \\
    Short & Transformer & Ball+Racket & Concat & 60.2 & 89.1 & 39.5 & 51.7 & 50.4 & 79.3 \\
    Short & Transformer & Ball+Racket & CrossAttn & 43.6 & 65.7 & 27.8 & 45.8 & 40.1 & 63.7 \\
    \midrule
    \multicolumn{10}{l}{\textit{Long Trajectory Prediction (History=80, Future=20)}} \\
    \midrule
    Long & LSTM & Ball & - & 113.9 & 184.3 & 62.5 & 108.7 & 118.7 & 194.7 \\
    Long & LSTM & Ball+Racket & Concat & 139.9 & 198.9 & 76.8 & 125.0 & 134.5 & 203.3 \\
    Long & LSTM & Ball+Racket & CrossAttn & \textbf{101.3} & \textbf{161.3} & \textbf{55.5} & \textbf{94.7} & \textbf{114.6} & \textbf{187.6} \\
    Long & Transformer & Ball & - & 145.3 & 207.5 & 89.9 & 144.8 & 142.7 & 228.3 \\
    Long & Transformer & Ball+Racket & Concat & 152.8 & 218.8 & 107.9 & 177.8 & 146.8 & 239.7 \\
    Long & Transformer & Ball+Racket & CrossAttn & 127.3 & 195.6 & 74.8 & 118.4 & 122.5 & 200.2 \\
    \bottomrule
  \end{tabular}
  \caption{Trajectory Prediction Experimental Results on RacketVision. The table shows the performance of different models and fusion methods under Short (History=20, Future=5) and Long (History=80, Future=20) prediction settings. The best result for each metric within each setting is highlighted in \textbf{bold}.}
  \label{tab:trajectory_prediction_results}
\end{table*}

\textbf{Evaluation Metrics.} We use two metrics for trajectory prediction:
\begin{itemize}
    \item \textit{Average Displacement Error (ADE)}: The average Euclidean distance between the predicted and ground-truth ball positions over all \(M\) future frames. The error is measured in pixels, assuming a resolution of 1920x1080.
    \item \textit{Final Displacement Error (FDE)}: The Euclidean distance between the predicted and ground-truth ball positions at the final frame (\(t+M\)).
\end{itemize}

\noindent
\textbf{Baseline Models.} We evaluate two backbone architectures, LSTM and Transformer, under three different input and fusion settings. This allows us to analyze not only the performance of the backbones but also the effectiveness of different multi-modal fusion strategies.

\begin{itemize}
    \item \textbf{Backbones}:
    \begin{itemize}
        \item \textit{LSTM}~\cite{hochreiter1997long}: A 2-layer recurrent neural network that models temporal dependencies sequentially through a stateful, recurrent mechanism.
        \item \textit{Transformer}~\cite{vaswani2017attention}: A 4-layer encoder-only model that captures global dependencies in the sequence in parallel via its self-attention mechanism.
    \end{itemize}
    \item \textbf{Input and Fusion Methods}:
    \begin{itemize}
        \item \textit{Ball-Only}: A strong unimodal baseline where the model only receives the historical ball coordinates as input.
        \item \textit{Concat Fusion}: A naive multi-modal baseline. The embeddings of the ball coordinates and racket poses are concatenated along the feature dimension before being fed into the backbone. This method treats all features equally at every time step.
        \item \textit{Cross-Attention Fusion}: The ball trajectory sequence acts as the \textit{Query} and the racket pose sequence acts as the \textit{Key} and \textit{Value}. This allows the model to dynamically weigh and extract the most relevant racket pose information for each time step of the ball's trajectory, effectively filtering noise and focusing on critical events like impacts.
    \end{itemize}
\end{itemize}

\noindent
\textbf{Experimental Results.} 
Table~\ref{tab:trajectory_prediction_results} summarizes the performance of baselines under the short and long trajectory prediction setting. Figure~\ref{fig:traj_pred_compare} provides a visual comparison of the trajectory prediction results between best models with ball-only input and ball+racket input. 

\begin{itemize}
    \item \textbf{Naive Fusion Degrades Performance.} As shown in Table~\ref{tab:trajectory_prediction_results}, simply concatenating racket pose features consistently leads to worse performance than the ball-only baseline for both LSTM and Transformer backbones. This is likely because a large portion of trajectory samples in our dataset capture the ball in mid-flight, where racket information is absent or irrelevant. The Concat method indiscriminately fuses this noisy or uninformative data, which hinders the model's ability to learn the primary trajectory dynamics.

    \item \textbf{Cross-Attention Excels at Predicting Critical Events.} The LSTM model equipped with Cross-Attention fusion is the best-performing model overall, demonstrating that racket information is highly valuable when integrated intelligently. The qualitative results in Figure~\ref{fig:traj_pred_compare} reveal precisely why this method is effective. For the Tennis sample ((a) vs. (b)), the Cross-Attention model leverages the racket's position to more accurately predict the trajectory's turning point. Similarly, for the Badminton sample ((c) vs. (d)), the model correctly infers the post-hit direction from the racket's pose. This shows that the Cross-Attention mechanism successfully learns to identify and heavily weigh racket features during critical "event" frames (i.e., hits), which are decisive for the subsequent trajectory. 

    \item \textbf{The Nature of Trajectory Data Explains Overall Gains.} While Cross-Attention provides a clear advantage during player-ball interactions, the overall statistical improvement in ADE/FDE over the strong ball-only baseline is noticeable but not dramatic. In Short Badminton, ADE improves from 37.5 to 37.0. This can be attributed to the dataset's composition: many samples, especially in the short-trajectory setting, consist entirely of the ball in flight, where no informative racket interaction occurs. In these common cases, the Cross-Attention model correctly learns to ignore the racket modality, effectively behaving like the ball-only model. 
\end{itemize}

\section{Conclusion}
\label{sec:conclusion}

In this work, we introduced RacketVision, a large-scale, multi-sport benchmark designed to advance sports analytics. By providing the first large-scale dataset with detailed annotations for both ball position and racket pose, we formulated three interconnected tasks—ball tracking, racket pose estimation, and ball trajectory prediction—to address key challenges in perception and motion forecasting.

Through extensive evaluation, we not only established strong performance benchmarks but also uncovered a critical insight into multi-modal fusion for trajectory prediction. We demonstrated that naively incorporating racket pose data via simple concatenation was detrimental to performance. However, a Cross-Attention architecture successfully unlocked the value of this contextual information, reversing the performance degradation and ultimately surpassing strong unimodal baselines. This key finding definitively demonstrates the dual importance of both the novel racket pose data and the advanced fusion architecture required to leverage it.

\section{Acknowledgments}
The work is supported by Shanghai Artificial Intelligence Laboratory.

\bibliography{aaai2026}

\end{document}